\documentclass[pdflatex,sn-mathphys-num,iicol]{sn-jnl}% Style for submissions to Nature Portfolio journals
\usepackage{makecell}
\usepackage{graphicx}%
\usepackage{multirow}%
\usepackage{amsmath,amssymb,amsfonts}%
\usepackage{amsthm}%
\usepackage{mathrsfs}%
\usepackage[title]{appendix}%
\usepackage{xcolor}%
\usepackage{textcomp}%
\usepackage{manyfoot}%
\usepackage{booktabs}%
\usepackage{algorithm}%
\usepackage{algorithmicx}%
\usepackage{algpseudocode}%
\usepackage{listings}%
\theoremstyle{thmstyleone}%
\theoremstyle{thmstyletwo}%

\theoremstyle{thmstylethree}%

\begin{document}

\title[Effects of Robotic Touch on Older Adults During Walking Guidance by a Humanoid Robot]{Effects of Robotic Touch on Older Users During Walking Guidance by a Humanoid Robot}


\author[1]{\fnm{Leonie} \sur{Leven}}\email{leonieleven@gmx.de}

\author*[1]{\fnm{Marko} \sur{Ackermann}}\email{marko.ackermann@kit.edu}

\author[2]{\fnm{Christian} \sur{Werner}}\email{christian.werner@med.uni-heidelberg.de}

\author[3]{\fnm{Melina} \sur{Schmetterer}}\email{melina.schmetterer@gmail.com}

\author[2]{\fnm{Theresa} \sur{Buchner}}\email{theresa.buchner@med.uni-heidelberg.de}

\author[3]{\fnm{Monika} \sur{Eckstein}}\email{monika.eckstein@med.uni-heidelberg.de}

\author[1,4]{\fnm{Katja} \sur{Mombaur}}\email{katja.mombaur@kit.edu}


\affil[1]{\orgdiv{Institute for Anthropomatics and Robotics}, \orgname{Karlsruhe Institute of Technology (KIT)}, \orgaddress{\state{Karlsruhe}, \country{Germany}}}

\affil[2]{\orgdiv{Geriatric Center, Medical Faculty Heidelberg}, \orgname{Heidelberg University}, \orgaddress{\state{Heidelberg}, \country{Germany}}}

\affil[3]{\orgdiv{Institute of Medical Psychology}, \orgname{University Hospital Heidelberg, and Ruprecht-Karls University Heidelberg}, \orgaddress{\state{Heidelberg}, \country{Germany}}}

\affil[4]{\orgdiv{Systems Design Engineering Department}, \orgname{University of Waterloo}, \orgaddress{\street{200 University Avenue West}, \city{Waterloo}, \postcode{N2L 3G1}, \state{Ontario}, \country{Canada}}}

%%==================================%%
%% abstract %%
%%==================================%%
\abstract{The shortage of healthcare staff is a challenge in geriatric care. To address this, robots can be integrated into care settings to provide assistance and emotional support. A promising application is walking guidance, particularly benefiting older adults as navigation skills deteriorate with aging. As walking guidance involves direct contact, the aim of this study is to understand how older adults perceive and respond to different touch modes during guided walking. 24 older adults (68 - 88 yrs.) walked four times a ten-meter trajectory guided by the robot TIAGo Pro in four contact conditions: no physical contact (NC); physical contact through holding the robot's wrist with the hand (HH); physical interaction through linking arms with the robot (LA); and physical contact through resting the forearm on the robots forearm (FC). A multimodal assessment approach included electrocardiogram, electrodermal activity, contact force, distance to robot, and questionnaires. Physiological results reveal a slight increase in stress levels during robot interaction. Behavioural and subjective measures, however, show overall acceptance of robotic touch. The two conditions corresponding to larger interaction forces (HH and FC) were associated with lower relative distances between participant and robot, indicating a higher trust and confidence.  Questionnaire responses supported these findings, evidencing greater perceived safety, trust and comfort in these conditions. This study provides insights for the design of robotic walking guidance assistance, indicating that gentle, stable touch is preferred by older adults in comparison to contactless interaction.}


\maketitle

\section{Introduction}\label{sec1}

Staff shortages in geriatric care represent a significant challenge in modern healthcare.
Already in $2034$ there will be a gap of $352.000$ missing caregivers in Germany \cite{statistisches_bundesamt_statistischer_2024}. To mitigate this problem, the possibility of using robotic assistants in healthcare facilities has gained increasing attention. Robots can support caregivers by taking over repetitive or demanding tasks, such as serving meals \cite{odabasi_refilling_2022}, or motivating residents to engage in physical and cognitive activities \cite{carros_exploring_2020}.

Beyond these contact-free applications, robots can also play an important role in providing physical and emotional support. Many physically assistive applications require direct contact between human and robot to transmit force and perform tasks effectively. Physical human contact has been shown to have a calming and stress-reducing effect \cite{morrison_keep_2016} and plays also a significant role in communication \cite{hertenstein_touch_2006}.
Therefore, it is reasonable to consider touch as a relevant factor in human–robot interaction.
A foundational review by Eckstein et al.\ \cite{eckstein_calming_2020} examined touch in human–human, human–animal, and human–robot contexts. The authors emphasized that research on robotic touch is still in its early stages but holds great potential, particularly in healthcare settings.

To address the emotional needs of patients when caregivers are overburdened, emotional support robots such as Paro have been developed \cite{wada_robot_2006}. Research on physical human-robot interaction has explored various aspects, including social touch \cite{willemse_social_2019, fitter_how_2020} and physical guidance \cite{nakane_development_2023, hieida_walking_2020}. However, most studies have focused on general adult populations rather than older adults. The perception of older adults was also researched in the literature \cite{wu_acceptance_2014}, however those studies did not involve physical human-robot interaction. One exception is the study by Piezzo and Suzuki \cite{piezzo_design_2016}, who investigated robotic touch during guidance with older adults. However, only one of four participants chose to physically interact with the robot.

One promising application to address the lack of caregivers is the use of robots to guide patients. Older adults could benefit greatly from such assistance, as navigation skills decline with aging \cite{moffat2009aging} and cognitive impairment, which affects particularly geriatric patients in rehabilitation centers, hospitals, and nursing homes \cite{Boustani2010,bjork2016exploring}. However, the general acceptance of this type of assistive robots, and the influence of different robotic touch modes on how humans perceive and respond to robotic systems are still not fully understood. In this context, the present study investigates how older adults respond to different robotic touch modes during guided physical interaction in terms of acceptance, trust, stress levels, and subjective perception. This study is complemented by a more in depth investigation of the influence of individual differences in social touch attitudes on older adults' experiences with robot-assisted walking guidance including touching the robot \cite{eckstein2026individual}.

This paper is structured as follows: Section~\ref{ch:related_work} provides insight into related work on robotic guidance, geriatric robot perception, and physical human-robot interaction.
Section~\ref{ch:technical_methods} describes the hardware used, the study protocol as well as the acquisition and treatment of the data. The results are presented in Section~\ref{ch:results}, and their evaluation are discussed in Section~\ref{ch:evaluation}.

\section{Related Work}\label{ch:related_work}

Robots which assist older adults in different activities and tasks have been investigated for a long time. Asgharian et al. \cite{asgharian_review_2022} summarized existing mobile service robots which were used to enhance the well-being of older adults and decrease the workload of caregivers. Starting from $1999$, Care-O-bot \cite{schraft_care-o-botsup_1998} was developed to assist in home environments. Reiser et al. \cite{reisercareobot2009} employed Care-O-bot 3 to test the applicability and acceptance of a robot serving water.

Wu et al. \cite{wu_acceptance_2014} conducted a study with $11$ older adults in a living lab setting, with sessions lasting one hour once per week over a period of four weeks. Participants engaged with the robot Kompa\"{i} using either the touchscreen or voice commands. Possible interactions included, for example, checking the calendar, adding appointments, or playing cognitive games. A robot-acceptance questionnaire was used to evaluate the perception of the participants after one and four weeks. Although they found the interaction to be easy, enjoyable, and non-threatening, they see no advantage in using the robot in everyday life. Their negative attitudes and acceptance toward robots also did not change after the experiences.

A more recently conducted study in a residential care home used the robot Pepper to identify potential enablers and barriers of using a robot-based assistance \cite{carros_exploring_2020}. In every session, there was a motivational phase at the beginning, a following physical exercise phase in which Pepper demonstrate exercise programs to the participants, and a cognitive phase afterwards. In the cognitive phase, games which are designed to promote memory, reaction or concentration were played. The interview questions at the end showed that most of them enjoyed interacting with Pepper, even though they were skeptical or even fearful at first.

Regarding the specific application of robotic guidance, Yoshimitsu et al. \cite{yoshimitsu_exploring_2024} performed a pilot study with the SEED platform robot, where participants were guided from the reception to goal destinations without physical contact. The results generally showed that people were satisfied with the guidance robot. There are also studies focusing on guiding older adults. Montemerlo et al. \cite{montemerlo_experiences_2002}, for example, used Pearl \cite{pollack_pearl_2002} to guide in an assisted living facility. The focus of that work was on evaluating the applicability of a partially observable Markov decision process, without considering the human perception of robotic guidance.

Willemse and van Erp \cite{willemse_social_2019} explored the calming effects of robotic touch while participants watched a scary movie. They additionally examined whether the presence of a prior bond between the robot and the human, established through earlier social interaction, had an influence. Independent if there was a prior bond to the robot, the results showed that robotic touch reduces physiological stress and increases the perceived intimacy of the human-robot bond. An approach to engage with robots in a lighthearted manner is presented in the study by Fitter and Kuchenbecker \cite{fitter_how_2020}. The authors investigated participants’ feelings while clapping hands with a robot, varying its facial reactivity, physical reactivity, arm stiffness, and clapping tempo. The results showed that increased facial reactivity made the robot appear more pleasant, while higher arm stiffness made it appear safer and less dominant. Faster tempos increased dominance.

The combination of the robotic guidance with the focus of physical human-robot interaction was also investigated in literature \cite{nakane_development_2023, hieida_walking_2020, piezzo_design_2016}.
Nakane et al. \cite{nakane_development_2023} developed a five-fingered robotic hand with the purpose of holding hands with a person. The authors attached the hand on a guidance robot and recorded guidance with a person two times, with and without holding hands. Participants only watched those two videos and subsequently completed a questionnaire. The results indicate that the perceived sense of security, regarding comfort and performance, was higher in the hand-holding condition. However, in the no contact condition participants rated controllability and perceived robot-likeness higher than in the hand-holding condition. 

Hieida et al. \cite{hieida_walking_2020} investigated if walking hand-in-hand with a robot has an effect on building a relationship between children and a robot. They designed and performed a study in which children walked while holding hands with a robot. To evaluate the interaction, the researchers measured the distance between the robot and the child, observed the direction of gaze, collected electrocardiogram (ECG) data, and administered questionnaires. The results suggest a positive effect of walking hand-in-hand on the building of a relationship between child and robot.

Piezzo and Suzuki \cite{piezzo_design_2016} investigated the reactions of older adults to physical human-robot interaction during walking in a nursing home, using the humanoid robot Pepper. To avoid overexerting in this population, they were asked to choose one of three contact variants: no contact, touching one shoulder of the robot while walking side-by-side, and touching the two shoulders of the robot while walking behind the robot. Only one out of the total of four participants felt confident enough to walk with the robot with side-by-side contact, the other ones preferred walking without contact. For evaluation, they measured the step length, step width and walking speed and asked the participants to fill out a shortened version of the Intrinsic Motivation Inventory questionnaire \cite{ryan1982control}. The reduced number of participants and the fact that only one participant walked with physical contact are important limitations.

\section{Methods}\label{ch:technical_methods}

This study investigates how older adults respond to different modes of robotic touch during guided physical interaction in terms of acceptance/trust, stress levels, and subjective perception. For this purpose, the robot TIAGo Pro by PAL Robotics \cite{palrobotics_tiagopro_2024} was used to guide the participants with four different  contact conditions: \emph{i}) no physical contact, in which the participant walked side by side with the robot; \emph{ii}) physical contact through holding the robot's wrist with the hand; \emph{iii}) physical interaction through linking arms with the robot, and \emph{iv}) physical interaction through resting the forearm on the robots forearm link. For the evaluation, a multimodal sensing setup was used. This section describes the robotic platform, the equipment and sensors used, the experimental setup and protocol, the evaluation metrics, and the data processing and statistical analysis.

\subsection{Robotic Platform}\label{se:robotic_platform}
The TIAGo Pro robot by PAL Robotics was selected as the robotic platform for this study, as it is specifically designed to operate safely and efficiently alongside humans.
Its series elastic actuators in the arms reduce the risk of injury during human–robot interaction while also allowing torque estimation at the joint level at a sampling rate of $50$ Hz.
The omnidirectional base provides increased stability compared to legged robots, although it is restricted to level-access environments. The robot’s base is equipped with two laser range finders that provide a $360$° field of view with a range of up to $25$ meters. It publishes data at a sampling rate of $15$ Hz. Additionally, its built-in RGB-D camera can be used to perform tasks such as human detection, and skeleton tracking. Both sensors are essential for this study, as they enable measurement of the participant’s position relative to the robot.  To communicate with the participants, the German voice ``Klaus'' of the text-to-speech engine from the Acapela Group\footnote{\url{https://www.acapela-group.com}}  was selected.

\subsection{Equipment} \label{se:equipment}
Additional sensors were required for the remaining evaluation metrics.
These included external cameras and sensors to record ECG and EDA physiological signals.
Three Sony RX0 II cameras connected to a camera control box for synchronization were used to capture the overall scene. To capture the ECG during the experiments, the EcgMove 4 by movisens\footnote{\url{https://www.movisens.com/de/produkte/ekg-sensor/}} was used. The sensor acquires a single channel ECG signal using disposable electrodes. The sampling rate for ECG acquisition was $1024$ Hz. The EdaMove 4 by movisens\footnote{\url{https://www.movisens.com/de/produkte/eda-sensor/}} was used to measure skin conductance. It records a single channel EDA signal using two Ag/AgCl electrodes. An exosomatic measurement method is used, with a constant DC voltage of $0.5$ V applied to the skin. The EDA signal was acquired at a sampling rate of $32$ Hz.

\subsection{Experiment Setup}\label{se:setup}
\begin{figure}
  \centering 
  \includegraphics[width=\linewidth]{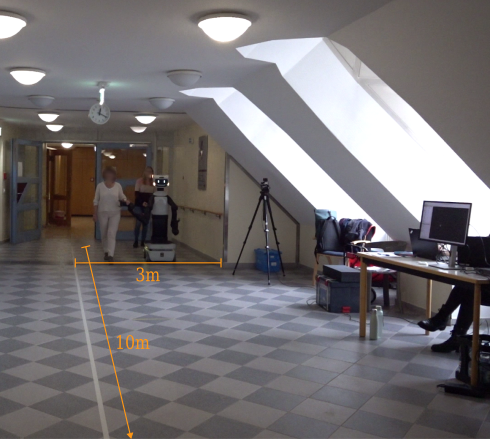}
  \caption{The experiment setup, with a straight, $10$~m-long path with an approximate width of $3$~m} 
  \label{fig:setup}
\end{figure}

The study took place on a less frequented floor of the Agaplesion Bethanien Hospital Heidelberg. Figure~\ref{fig:setup} shows the experimental setup. The path is 10~m long, and has an approximate width of $3$~m, with the experiment area marked with tape. Two cameras were placed at approximately $3$~m and $7$~m from the start of the path to capture the participant's face. An additional camera was positioned at the start of the path to record the entire experiment, enabling later review.

\subsection{Participants}\label{se:participants}
Participants were recruited from pools of older adults who had participated in previous studies conducted by the research department at the Geriatric Center (e.g., from mailing lists of individuals interested in research topics). The inclusion criteria were healthy, non-frail adults over $65$ years of age with no cognitive impairments (Mini-Mental State Examination \cite{folstein_mini-mental_1975} $\ge$ 25). Participants were physically fit, as indicated by a mean usual gait speed in a $4$~m gait test of $1.16$ m/s (SD $0.24$ m/s).

Participants were required to be able to walk independently for at least $20$ meters, four times consecutively. They were also asked to refrain from smoking or consuming alcohol and caffeine for at least two hours prior to their appointment to avoid influencing the biosignal measurements. In total, $24$ participants (3 male) were recruited, with an average age of $75.8$ years (SD $5.2$ years). A total of $54.2\%$ of participants had prior experience with robots, $69.2\%$ of whom had previously participated in a robotics study. $45.8\%$ had never interacted with a robot before. All participants provided their informed consent at the beginning of the study and participated in the study voluntarily.

\subsection{Experiment Procedure} \label{se:procedure}

\begin{figure*}
  \centering 
  \includegraphics[width=0.24\textwidth]{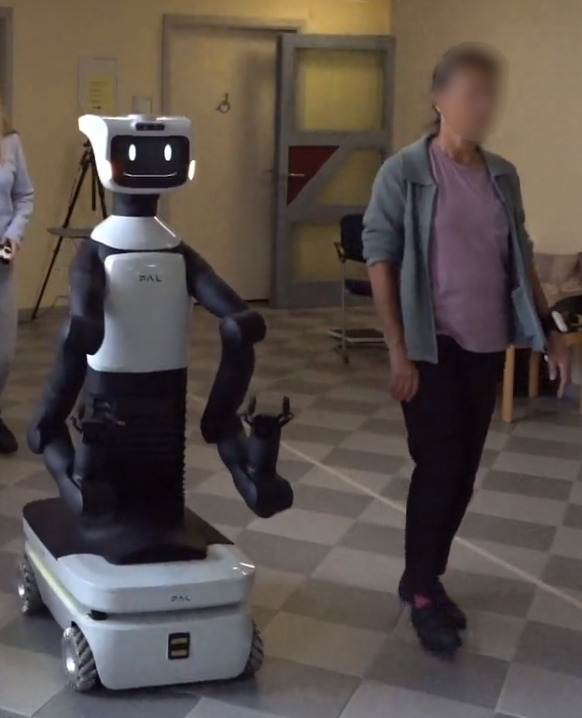}
  \includegraphics[width=0.24\textwidth]{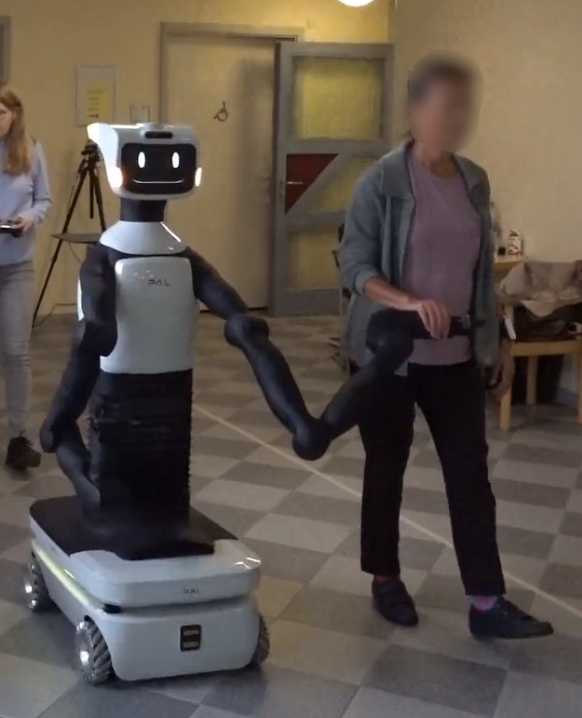}
  \includegraphics[width=0.24\textwidth]{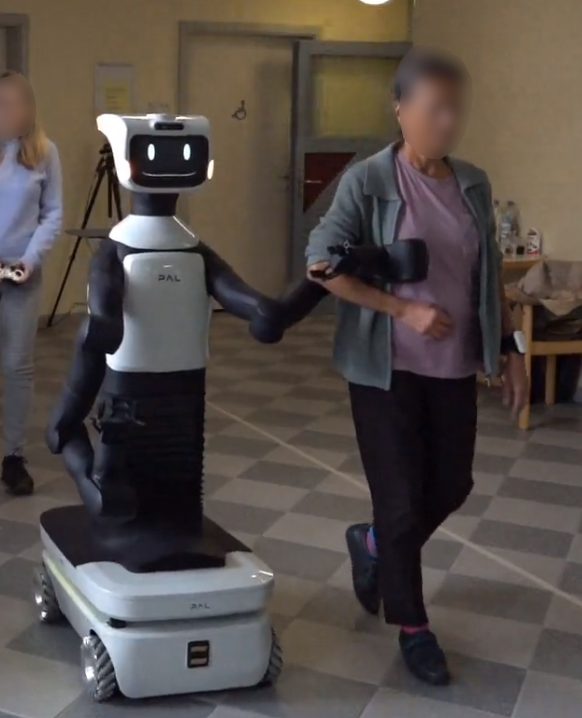} 
  \includegraphics[width=0.24\textwidth]{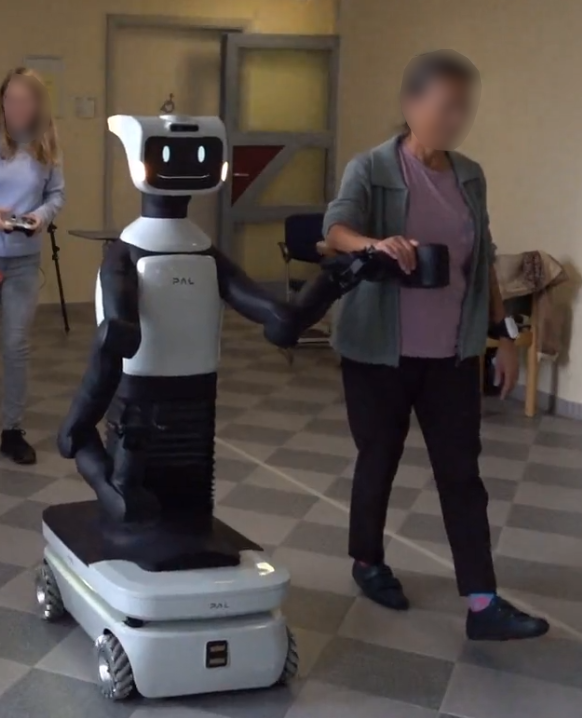}
  \caption{Walking with the robot under the four different contact conditions (from left to right): (\emph{i}) No Contact (NC); (\emph{ii}) Holding Hands (HH); (\emph{iii}) Linking Arms (LA); and (\emph{iv}) Full Forearm Contact (FC)} 
  \label{fig:four_conditions}
\end{figure*}

Two sensors were attached: one on the hand to measure the EDA and one on the lower left chest to record the ECG signal. Subsequently, participants completed questionnaires regarding their health and technical background, as well as the Negative Attitudes toward Robots Scale (NARS) \citep{nomura_measurement_2006}, the Computer Anxiety Trait Scale \cite{gaudron_assessing_2002}, and the Social Touch Questionnaire \cite{wilhelm_social_2018}.

Before the robot interaction began, participants were asked to walk the twenty-meter distance of one repetition without the robot at their normal gait speed. Their walking speed was measured over the initial four-meter segment, which is usual in clinical assessments \cite{karpman_measuring_2014} and is used to measure functional capacity. After returning, the robot introduced itself, and the interaction phase commenced.

The participants completed four trials with different touch conditions, respectively, each consisting of two repetitions. In each repetition, the robot travels ten meters straight ahead, then turns and returns along the same path. The four different walking conditions were presented in a randomized order for each participant. To ensure safety, the robot was remotely operated by the experimenter, who walked approximately two meters behind the robot and the participant (see Figure~\ref{fig:four_conditions}). The robot’s speed was controlled by the experimenter according to the participant's response but was limited to a maximum of $0.5$ m/s to maintain safe operation, as higher velocities made manual control more difficult. At the beginning of each trial, the robot positions itself in front of the participants and detects their body position using the body tracking module and the RGB-D camera. This is needed to connect the human's position information to the point cloud of the laser scanner. It then approaches the participants and extends its arm toward them, depending on the assigned condition. The four conditions are as follows (see example pictures in Figure~\ref{fig:four_conditions}):
\begin{itemize}
    \item  \textbf{No Contact (NC)}:
The robot waits a few seconds before saying ``Let's get going together now''. It then moves along the predefined path without touching the participant.

    \item \textbf{Holding Hands (HH)}:
The robot extends its gripper and says ``Please place your hand on my wrist''.
It waits until the participant holds its wrist link.

    \item \textbf{Linking Arms (LA)}:
The robot offers its arm and says ``Please wrap your arm in my arm''.
It waits until the participant links arms with the robot.

    \item \textbf{Full Forearm Contact (FC)}:
The robot offers its arm and says ``Please put your arm on mine''.
It waits until the participant rests their forearm on its forearm link.
\end{itemize}

In all contact conditions, after waiting until the contact is established, the robot says ``Let's get going together now''. It then moves along the predefined path. For all conditions, shortly before reaching the end of each path in one direction, the robot says ``Attention, we're about to stop. It would be nice if you could let go of me, then we can each turn around and walk back''. When the trial is completed, the robot thanks the participant saying ``Thank you very much, we are done with this trial''. All the lines were originally spoken by the robot in German and were translated to English here. In the LA and FC conditions, participants were given the opportunity to adjust the robot’s height to their preference.

After each trial, a brief interview-based questionnaire was administered, consisting of seven items addressing perceived safety, discomfort, tension, and trust during the guidance, as well as comfort, convenience, and naturalness of the physical contact. In addition, the short version of the Positive and Negative Affect Schedule (PANAS) questionnaire \citep{watson_development_1988} was included. Participants were also asked an open-ended question about their current mood  and to provide a rating between $1$ (``very bad'') and $10$ (``very good'') evaluating that specific condition. The entire robot interaction concluded with the robot saying ``Thank you very much for allowing me to accompany you. I hope you enjoyed it. Goodbye and see you next time''.

\begin{figure*}
  \centering 
  \includegraphics[width=\linewidth]{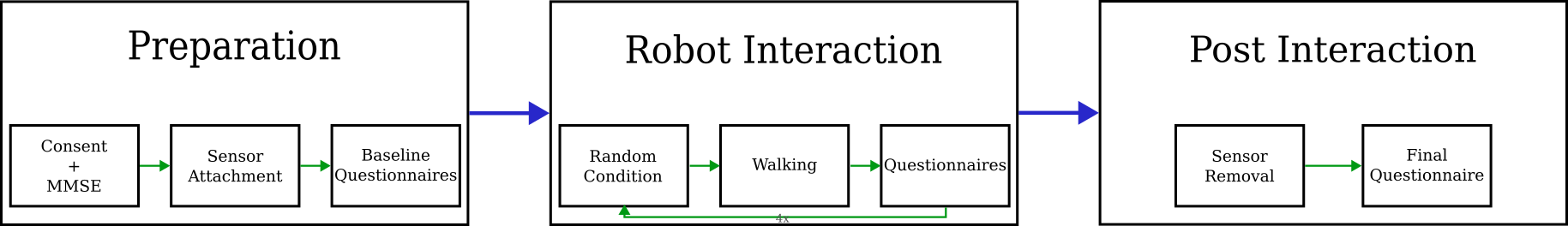}
  \caption{Flowchart of the experimental protocol. The robot interaction block was repeated four times, for each contact condition in a randomized order} 
  \label{fig:procedure}
\end{figure*}

Finally, the sensors were removed, and participants were asked to complete a final questionnaire consisting of the NARS and an open-ended comment question. The entire protocol took approximately one hour per participant and is illustrated in Figure~\ref{fig:procedure}.

% \begin{figure*}
%   \centering 
%   \includegraphics[width=\linewidth]{Fig3_new.png}
%   \caption{Flowchart of the experimental protocol. The robot interaction block was repeated four times, for each contact condition in a randomized order.} 
%   \label{fig:procedure}
% \end{figure*}

\subsection{Evaluation} \label{se:evaluation_data}
For the evaluation, contact forces measure contact intensity, and average distance between the robot and the participant indicates trust and acceptance. ECG and EDA are used to assess levels of stress (anxiety, fear), and questionnaires are used to inform subjective perception. 

\subsubsection{Stress Levels} \label{se:ev_biosensor}
ECG and EDA data serve as indicators of stress \cite{kreibig_autonomic_2010, boucsein_electrodermal_2012}. Moreover, as these parameters are not under voluntary control, they provide a more objective insight into participants' perception.

For the evaluation, the following ECG parameters were used:
\begin{itemize}
    \item \textbf{Heart rate (\textit{HR}, 1/min)}: Mean value of the heart rate calculated over a $30$-second interval. Larger \textit{HR} indicates higher stress, anger, anxiety and/or fear \cite{kreibig_autonomic_2010}.
    \item \textbf{HRV parameter \textit{RMSSD} (ms)}: Mean value of the Root Mean Square of Successive Differences (\textit{RMSSD}) of beat intervals, computed over a sliding two-minutes time window updated every $30$ seconds, using the preceding two minutes of data. A lower \textit{RMSSD} indicates higher stress, anger, anxiety and/or fear \cite{kreibig_autonomic_2010}.
    \item \textbf{HRV parameter \textit{Pnn50} (\%)}: The percentage of normal-to-normal beat intervals greater than $50$ ms. The mean is calculated over a sliding two-minute time window updated every $30$ seconds, using the preceding two minutes of data. A lower \textit{Pnn50} indicates higher stress, anger, anxiety and/or fear \cite{kreibig_autonomic_2010}.
\end{itemize}

Using the EDA raw data, the following parameters were derived for the evaluation, where larger values indicate higher stress levels, anger, anxiety and/or fear \cite{kreibig_autonomic_2010}:
\begin{itemize}
    \item \textbf{Skin conductance level (\textit{SclMean}, $\mu$S)}: Mean value of the SCL over a ten-second interval.
    A low-pass filter was applied to the raw EDA data beforehand to remove noise. 
    \item \textbf{Number of skin conductance responses (\textit{ScrCount})}: Accumulated number of SCRs over a ten-second interval.
    \item \textbf{Mean SCR amplitude (\textit{ScrAmplitudesMean}, $\mu$S)}: Mean value of the amplitudes of the SCRs over a ten-second interval.
\end{itemize}

Both sensors were attached approximately ten minutes before the robot interaction to obtain baseline measurements. This means that the participants already wore the sensors while completing the initial set of questionnaires.

The ECG data were used to compare physiological response between the baseline phase and the entire robot interaction. ECG data were not analyzed on a condition-specific basis because of the limited trial durations ($2$-$3$~min), considering that \textit{HRV} parameters are calculated as an average over two-minute windows with updates every $30$ seconds and the \textit{HR} is calculated over a $30$-second interval.  

In contrast, the EDA data were used to compare the different touch conditions. Only time segments during which participants were walking with the robot were included, excluding the turning period and the subsequent interview sessions. All values were baseline-corrected to ensure comparability across participants. Because EDA values are dependent on the chosen baseline, two baselines were defined. The first baseline corresponded to a five-minute segment during the initial questionnaire session, representing a relaxed, seated state, similar to the ECG baseline. The second one was obtained during the initial gait speed measurement, representing walking without the robot and thus serving as a meaningful comparison condition. In order to get relative values, the mean baseline value was subtracted from each condition's mean for the parameters \textit{SclMean} and \textit{ScrAmplitudesMean}. For \textit{ScrCount}, instead of comparing the absolute numbers of SCRs, the rates were used. Accordingly, the baseline rate (number of SCRs divided by the duration) was subtracted from the rate measured for the different phases.

Subjects with incomplete data and outliers were removed prior to the subsequent statistical analysis.

\subsubsection{Intensity of Contact} \label{subse:force}
The force a person exerts on the robot can provide insights into the intensity of contact and the nature of the physical interaction, indicating pure guidance cues or the possibility of actual physical support. The interaction force, occurring at different locations or surfaces depending on the contact condition, is represented by the equivalent force applied at a point of interest. This force is estimated from the measured joint torques, according to the following procedure.   

The mean torque per joint was calculated  considering only periods in which there was contact between the participant and the robot. Data collected while the robot was not in contact with the participant, e.g.\ during turning, were disregarded. The left and right joints were considered separately, and data from the inactive arm were discarded. To compensate for gravitational effects and compute the force due to human-robot interaction only, for each condition one additional trial without human interaction was recorded to capture the joint torques due to gravity in the corresponding arm pose. The mean torque values from these control trials were subtracted from the corresponding mean torques obtained during the experimental trials.

The applied force was computed from the corresponding mean torques (after subtracting the contribution of gravity) using the pseudo-inverse of the Jacobian matrix for the point of interest. For this analysis, the segment origin expected to experience the highest force in each condition was selected as the point of interest for the Jacobian matrix computation: wrist link (segment 7) for HH, and forearm link (segment 5) for LA and FC. For NC, the intermediate robot link (segment $6$) was used, however, computed forces (after gravity compensation) were close to zero, as expected because there is no contact between robot and human in this condition. Finally, the mean force magnitude across all participants was calculated for each condition and arm segment and the values from the right and left sides were averaged.

\subsubsection{Interaction Intensity and Trust} \label{subse:distance}
We hypothesize that proximity is an indicator of the level or intensity of interaction, and that a smaller distance between the robot and the participant reflects higher trust and confidence. 

During the experiments, the robot continuously measured the distance and angle of the participant relative to itself. Although this could be achieved using the RGB-D camera and existing human detection algorithms, a robot that constantly faces the participant while walking appears unnatural. Therefore, it is more appropriate to rely on data from the laser scanners. The laser scanners produce a two dimensional point cloud at a height of $22$ cm parallel to the ground with a $360$° field of view up to $25$ m.

To determine the participant’s distance from the robot, several processing steps are applied to the point cloud, refer to Appendix~\ref{app:distance}. In summary, an axis-aligned bounding box relative to the robots frame is determined for the point clusters identified as associated with the participants legs. The distance between the robot and the participant is represented by the distance between the center of this bounding box and the origin of the robot's frame (located at middle of the robot's platform). The distance measurements were then averaged, excluding the data for turning periods, which did not involve physical guidance and contact.

\begin{table}
    \centering
    \caption{Closest and furthest possible distances for each condition while maintaining the corresponding contact conditions: NC - No Contact; HH - Hand Holding; LA - Linking Arms; FC - Full Forearm Contact}
    \begin{tabular}{lcc}
    \toprule
    \textbf{Condition} & \textbf{Lower Limit} & \textbf{Upper Limit}\\
    \midrule
    NC & 0.44 m & 1.50 m\\
    HH & 0.71 m & 1.04 m\\
    LA & 0.62 m & 0.75 m\\
    FC & 0.66 m & 0.95 m\\
    \bottomrule
    \end{tabular}
    \label{tab:distance_limits}
\end{table}

It is expected that the observed distances highly depend on the contact condition, and corresponding human and robot poses. HH generally allows the participant to maintain a larger distance to the robot compared to FC. In contrast, LA is the most restrictive, forcing the participant to remain close to the robot with a narrow range of possible distance variation. For this reason, a comparison of the absolute distances would be skewed and not necessarily representative of confidence or trust. To remediate this, a relative distance is defined, corresponding to the absolute distance normalized by the range of distances allowed by each condition, where $0$ represents a distance corresponding to the lower bound, and $1$ to a distance corresponding to the upper bound. The lower and upper bounds were determined for each contact condition (HH, LA, FC) by tests performed by two individuals while attempting to be as close and as far away as possible while still within the specified contact condition. The upper and lower limits obtained in this manner and used to normalize the distance and compute the relative distances are reported in Table~\ref{tab:distance_limits}. For the NC condition the upper limit was assumed to be the expected distance between the robot and the participant considering the lateral limits of the path (Fig.\ \ref{fig:setup}).

\subsubsection{Expression of Emotion} \label{subse:fer}
Expressing emotions through facial expressions is a very natural form of human communication, and is used in the literature to obtain the current mood of a human \cite{xiao_-road_2022, hoffmann_persuasive_2021}. Two cameras were placed in the experimental setup to allow for later automatic identification of facial emotion. Unfortunately,  during post-processing it became clear that the data was unreliable due to unfavorable image acquisition conditions. Inconsistent lighting, distance to cameras, and the downward gaze of the participants reduced image quality, preventing an accurate evaluation and leading to often wrongly labeled frames, even after trying different available software. For this reason, we decided not to further use the data in the evaluation.

\subsubsection{Subjective Perception} \label{subse:questionnaires}
When evaluating assistive technology, the subjective perception of the user is very important, alongside objective measurements. This is assessed in this study by a set of questionnaires.

After each trial an interview session was conducted. The first part consisted of seven questions concerning the participants' feelings during the interaction, rated on a 5-point Likert scale ($1$ - fully disagree, $5$ - fully agree), as well as an overall evaluation of the conditions on a scale from $1$ to $10$ (higher values indicating more positive evaluation):
\begin{enumerate}
    \item I felt safe while walking with the robot;
    \item I felt uncomfortable while walking with the robot;
    \item I was tense while walking with the robot;
    \item I trusted the robot while walking;
    \item Physical contact with the robot was pleasant;
    \item I felt comfortable during physical contact;
    \item The type of contact was natural;
    \item Overall evaluation.
\end{enumerate}

Additionally, participants answered an open-ended question on their current mood.

Next, the PANAS \citep{watson_development_1988} was conducted as an oral interview. The scales are often used in user studies to assess the emotional perception from positive and negative dimensions (e.g., used in \cite{willemse_social_2019, hoffmann_persuasive_2021, guo_how_2024}). In this study a subset of this questionnaire was used after each condition. We decided not to use the whole scale each time, as this would be excessively time consuming and exhausting, especially for older adults. Instead, we used the same selection of adjectives (``enthusiastic'', ``inspired'', ``afraid'', ``hostile'', ``interested'', ``nervous'') as in the related studies with the Navel robot \cite{yamamoto2026perceptionsocialrobotscommunication, mayer2026trust} (preprints). For the statistical analysis, the positive and the negative affects were considered separately. The mean positive affect score was calculated by first averaging each participant's ratings of the three positive adjectives, and then computing the overall mean of these participant averages across all participants. The variance is calculated across the mean values of the participants. The mean and variance of the negative affect was calculated in the same way.

Similar to the studies in \cite{willemse_social_2019} and \cite{zhou_tactile_2021}, the NARS \citep{nomura_measurement_2006} was used to evaluate if the interaction with the robot has any effects on the participants' attitudes. The items in the NARS are assigned to different sub-scales which consist of \textit{Negative Attitudes toward Situations and Interactions with Robots}, \textit{Negative Attitudes toward Social Influence of Robots}, and \textit{Negative Attitudes toward Emotions in Interaction with Robots}.

A more in-depth analysis of the influence of individual differences in  social touch on older adult's perception measured through these questionnaires is reported in a complementary paper \cite{eckstein2026individual}.

German versions of all questionnaires were used.

\subsection{Statistical Analysis} \label{se:statistics}
Not every participant contributed data to all parameters due to occasional sensor malfunctions or improperly attached devices, which led to distorted or missing signals.
Additionally, distance and force data were unavailable for some of the participants.
Thus, data from $14$ participants were included for ECG analyses, $24$ for EDA, $16$ for force, and $15$ for distance.

After the post-processing of the individual data and parameters as described in the previous section, normal distribution was tested using the Shapiro–Wilk test. If normal distribution is identified, a repeated-measures ANOVA is conducted. A subsequent post-hoc analysis using pairwise t-tests with Bonferroni correction was performed where appropriate. For the ECG data, as there are only two conditions which are compared, only the paired t-test was performed.

For the EDA parameters the Shapiro–Wilk test indicates that the resulting data were not normally distributed. Thus, the non-parametric Friedman test was applied to assess statistical significance. A subsequent post-hoc analysis using pairwise Wilcoxon tests with Bonferroni correction was conducted for the \textit{ScrAmplitudesMean} parameter under both baseline correction conditions. Also the data of the NARS questionnaire were not normally distributed. As there are only two conditions, before and after the interaction with the robot, the paired Wilcoxon test was directly applied.

\section{Results}\label{ch:results}
The increased \textit{HR} and decreased \textit{HRV} shown in Table~\ref{tab:descriptive} indicate a tendency toward a slight increase in stress during robotic interaction compared to the baseline phase. However, these differences were not significant, as evidenced by the t-test results in Table \ref{tab:statistics}.  

\begin{figure}
    \centering
    \includegraphics[width=.45\textwidth]{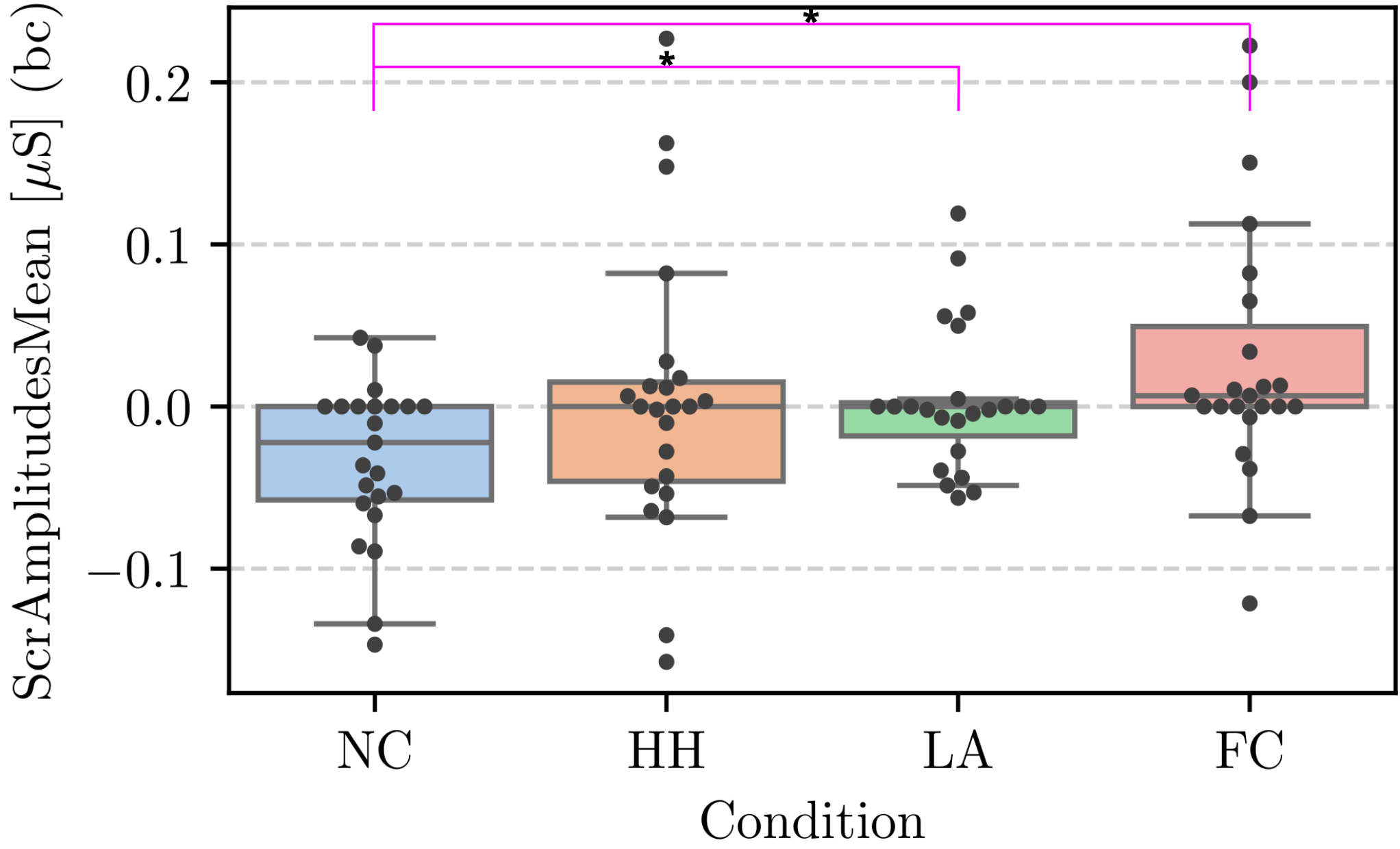}
    \caption{Box-plots of the baseline-corrected (bc) EDA parameter \textit{ScrAmplitudesMean} values after subtraction of the sitting baseline values during the robot interaction across the four different conditions: NC - No Contact; HH - Hand Holding; LA - Linking Arms; FC - Full Forearm Contact. * depicts $\text{p} < 0.05$.}
    \label{fig:eda_boxplot}
\end{figure}

The EDA analysis, used to compare stress levels among the different contact conditions, showed a significant effect for \textit{ScrAmplitesMean} (Table~\ref{tab:eda_friedman}), with slight increase in values for the contact conditions compared to the contactless condition (Figure~\ref{fig:eda_boxplot}). No significant effects were observed for \textit{ScrCount} or \textit{SclMean} (Table~\ref{tab:eda_friedman}). For the \textit{ScrAmplitesMean} parameter, the pairwise post-hoc analysis revealed that these increases were statistically significant only when the sitting baseline was used and for the pairs LA-NC and FC-NC (Table~\ref{tab:eda_scramplitudes_posthoc_gait}), indicating that the contact conditions LA and FC may be associated with slightly higher levels of stress compared to the contactless guiding condition.

\begin{table}
\centering
\caption{Descriptive statistics for \textit{HR}, \textit{HRV-RMSSD}, and \textit{HRV-Pnn50} before and during the robot interaction (section~\ref{se:evaluation_data})}
\begin{tabular}{lcc}
\hline
\textbf{Variable} & \textbf{before} & \textbf{during} \\
\hline
\textit{HR} (bpm)    & 78.33 & 79.03 \\
\textit{HRV-RMSSD} (ms)    & 14.59 & 14.54 \\
\textit{HRV-Pnn50}       & 0.83  & 0.75 \\
\hline
\end{tabular}
\label{tab:descriptive}
\end{table}

\begin{table}
\centering
\caption{Statistical test results for \textit{HR}, \textit{HRV-RMSSD}, and \textit{HRV-Pnn50} (section~\ref{se:evaluation_data})}
\begin{tabular}{lccc}
\hline
\textbf{Variable} & \textbf{t} & \textbf{p-value} & \textbf{Cohen’s d} \\
\hline
\textit{HR}     & -1.229 & 0.241 & 0.328 \\
\textit{HRV-RMSSD}    &  0.033 & 0.974 & -0.009 \\
\textit{HRV-Pnn50}  &  0.283 & 0.781 & -0.076 \\
\hline
\end{tabular}
\label{tab:statistics}
\end{table}

\begin{table}
\centering
\caption{Results of the Friedman test for all the EDA parameters (section~\ref{se:evaluation_data}) across conditions, corrected with two different baselines (gait and sitting), with significant differences ($\text{p} < 0.05$) highlighted in bold}
\begin{tabular}{lccc}
\toprule
\textbf{Parameter} & \textbf{Baseline} & \textbf{Statistic} & \textbf{p-value} \\
\midrule
\textit{SclMean} & Gait & 3.0800 & 0.3795 \\
            & Sitting & 4.0500 & 0.2561 \\
\textit{ScrCount} & Gait & 0.8305 & 0.8422 \\
             & Sitting & 0.6262 & 0.8904 \\
\textit{ScrAmplMean} & Gait & 8.1620 & \textbf{0.0428} \\
            & Sitting & 14.4612 & \textbf{0.0044} \\
\bottomrule
\end{tabular}
\label{tab:eda_friedman}
\end{table}

\begin{table}
\centering
\caption{Post-hoc tests (pairwise Wilcoxon with Bonferroni correction) for EDA parameter \textit{ScrAmplitudesMean} [$\mu$S] (baseline-corrected, sitting baseline), where: NC - No Contact; HH - Hand Holding; LA - Linking Arms; FC - Full Forearm Contact, and significant differences ($\text{p} < 0.05$) highlighted in bold}
\begin{tabular}{llccc}
\toprule
\textbf{A} & \textbf{B} & \textbf{Statistic} & \textbf{$p_\text{unc}$} & \textbf{$p_\text{corr}$} \\
\midrule
LA & FC & 55.000 & 0.1075 & 0.6448 \\
LA & HH & 94.000 & 0.6813 & 1.0000 \\
LA & NC & 18.000 & 0.0019 & \textbf{0.0117} \\
FC & HH & 54.000 & 0.0569 & 0.3415 \\
FC & NC & 20.000 & 0.0043 & \textbf{0.0260} \\
HH & NC & 57.000 & 0.0731 & 0.4388 \\
\bottomrule
\end{tabular}
\label{tab:eda_scramplitudes_posthoc_gait}
\end{table}

\begin{table}
\centering
\caption{Post-hoc comparisons of the force magnitudes between conditions (pairwise t-tests, Bonferroni-corrected), where: NC - No Contact; HH - Hand Holding; LA - Linking Arms; FC - Full Forearm Contact, with significant differences ($\text{p} < 0.05$) highlighted in bold}
\begin{tabular}{llccc}
\toprule
\textbf{A} & \textbf{B}  & \textbf{t} & \textbf{$p_\text{unc}$} & \textbf{$p_\text{corr}$} \\
\midrule
HH & FC  & -11.23 & 1.07e-08 & \textbf{6.41e-08} \\
HH & LA  & 2.85 & 1.22e-02 & 7.33e-02 \\
HH & NC  & 10.04 & 4.72e-08 & \textbf{2.83e-07} \\
FC & LA  & 8.44 & 4.42e-07 & \textbf{2.65e-06} \\
FC & NC  & 16.21 & 6.44e-11 & \textbf{3.86e-10} \\
LA & NC  & 3.97 & 1.22e-03 & \textbf{7.34e-03} \\
\bottomrule
\label{tab:t-test_force}
\end{tabular}
\end{table}

\begin{figure}
    \centering
    \includegraphics[width=.45\textwidth]{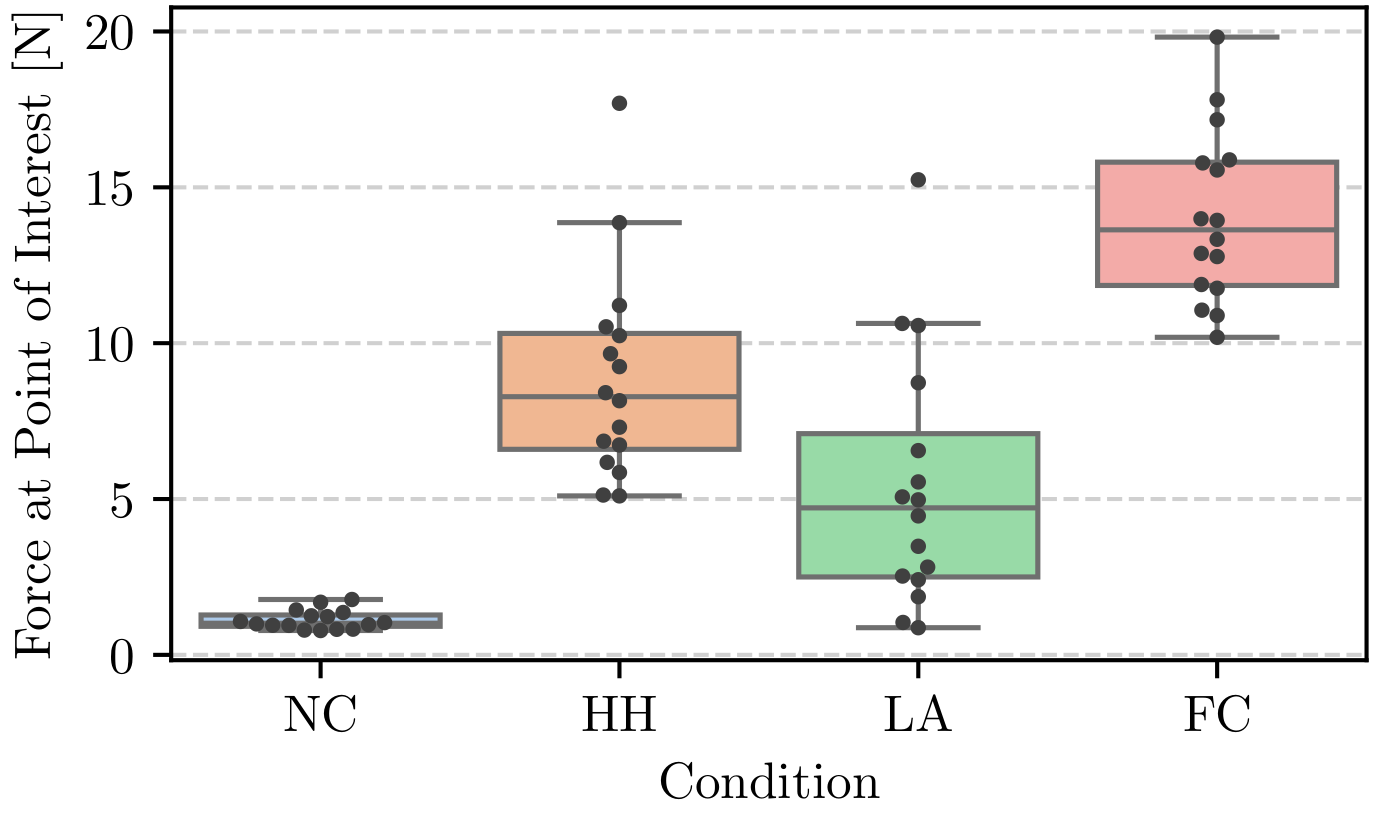}
    \caption{Box-plots of average equivalent force magnitudes for the different contact conditions: NC - No Contact; HH - Hand Holding; LA - Linking Arms; FC - Full Forearm Contact.}
    \label{fig:force_boxplot}
\end{figure}

As for the quantification of intensity of contact between the human and the robot, the magnitudes of the force for all the contact conditions are presented in Fig.\ \ref{fig:force_boxplot}, with median values of $1.2$~N for NC, $9.0$~N for HH, $5.9$~N for LA, and $14.3$~N for FC. The pairwise t-tests post-hoc comparisons (Bonferroni-corrected) of the force magnitudes between conditions revealed significant differences for all pairs except for HH-LA (Table \ref{tab:t-test_force}). 

\begin{figure}
    \centering
    \includegraphics[width=.45\textwidth]{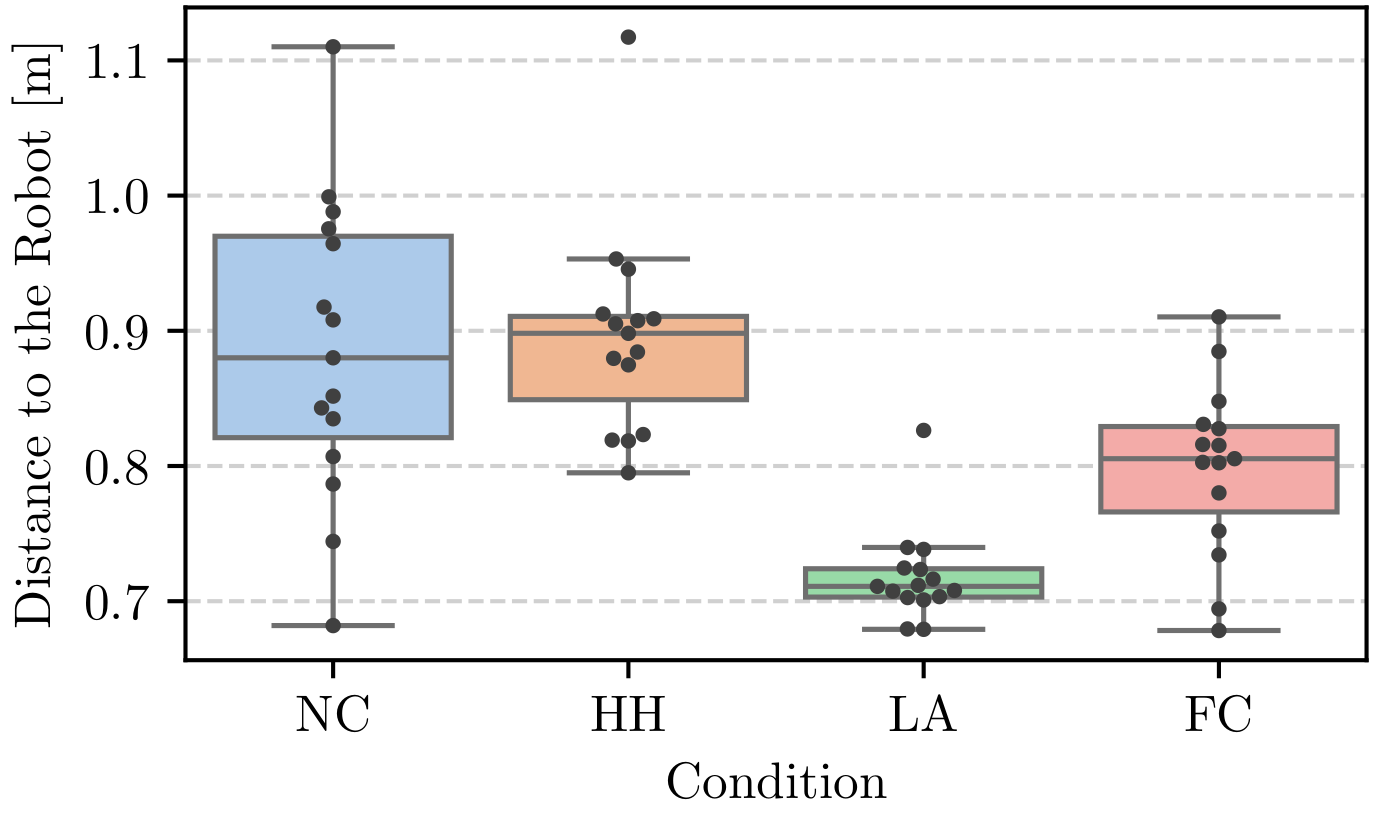}
    \caption{Box-plots of the absolute horizontal distance between the robot and the person (see Sect.~\ref{subse:distance}) for the different contact conditions: NC - No Contact; HH - Hand Holding; LA - Linking Arms; FC - Full Forearm Contact.}
    \label{fig:distance}
\vspace{10pt}
    \centering
    \includegraphics[width=.45\textwidth]{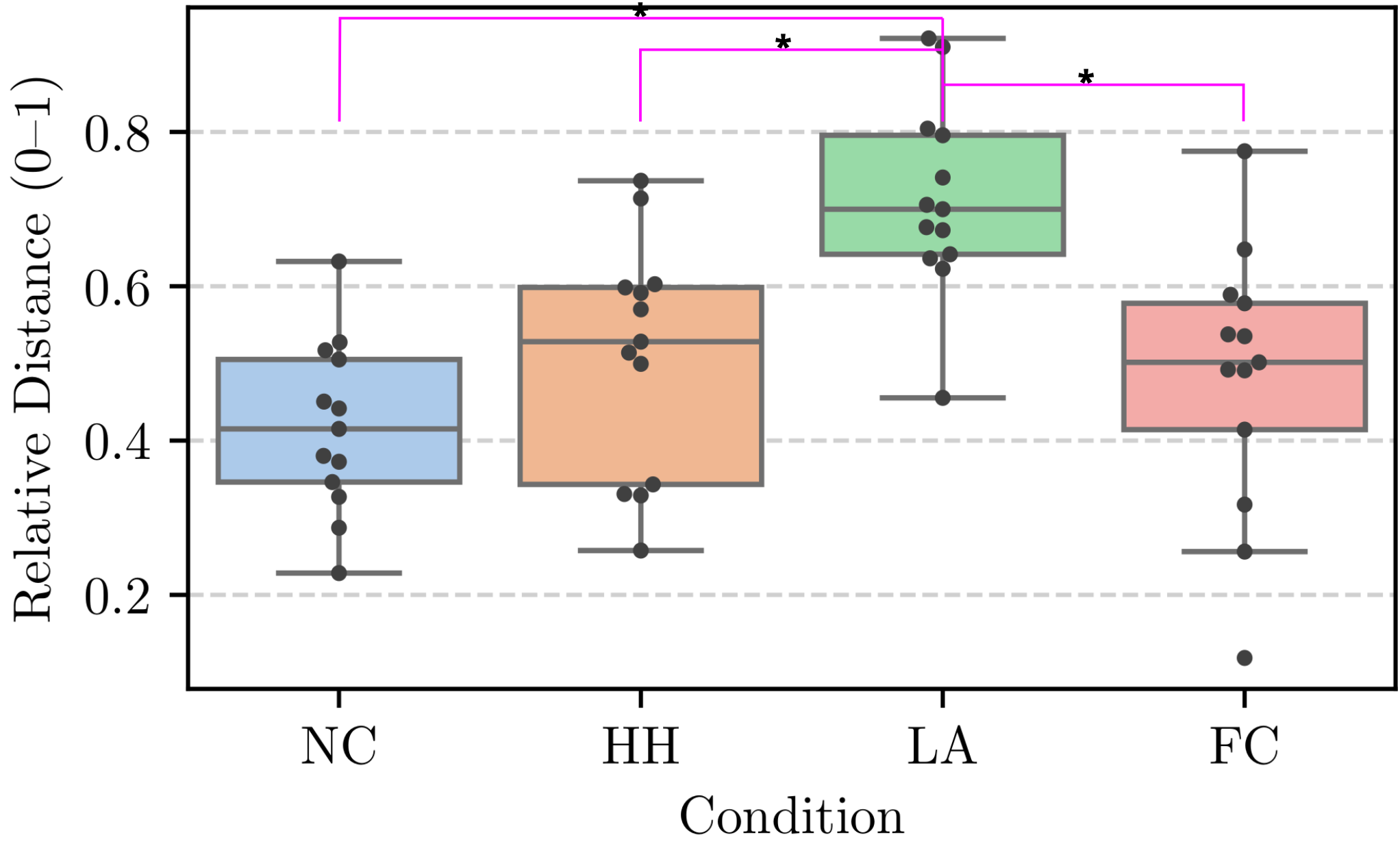}
    \caption{Box-plots of the relative (normalized) horizontal distance between the robot and the person (see Sect.~\ref{subse:distance}) for the different contact conditions: NC - No Contact; HH - Hand Holding; LA - Linking Arms; FC - Full Forearm Contact. * depicts $\text{p} < 0.01$.}
    \label{fig:distance_norm}
\end{figure}

The distance between the human and the robot was used to measure trust and confidence, and to quantify the intensity of the interaction. Figure~\ref{fig:distance} shows the absolute distances associated with each contact condition, which are influenced by the different robot arm poses, the contact modes, and how participants interact with robot. In LA and FC the robot's arm poses are the same, but the contact mode is different (linking arms with the robot in LA, and human forearm - robot forearm link for FC), with the LA requiring the participants to be much closer to the robot and offering limited range of lateral displacement. HH, on the other hand, allows the participant greater flexibility during interaction with the robot. In NC, participants chose to walk at a distance equivalent to the one in HH. 

To isolate the influence of the human response on the distance to the robot, the absolute distances were normalized by the range of lateral motions allowed by each contact condition resulting in the relative distances shown in Figure~\ref{fig:distance_norm}, with the repeated-measures ANOVA revealing a significant effect ($\text{p} < 0.01$). These results clearly show that while for NC, HH and FC participants chose to remain at about the middle of the range, with relative distances close to $0.5$, in the linking arm, participants chose to walk closer to the upper bound of the range, with a median relative distance of about $0.7$. The post-hoc analysis revealed that the difference of LA compared to all other conditions is statistically significant ($\text{p} < 0.01$). This indicates lower trust and confidence in the LA condition compared to the other guiding conditions.

\begin{table*}
    \small
    \centering
    \caption{Mean scores and standard deviations (SD) for the follow up questions after each contact condition: NC - No Contact; HH - Hand Holding; LA - Linking Arms; FC - Full Forearm Contact. The best score in each question is marked bold.}
    \begin{tabular}{lcccc}
    \toprule
        \textbf{Question} (best score) & \textbf{NC} & \textbf{HH} & \textbf{LA} & \textbf{FC} \\
        \midrule
        1. Safety (5) & 4.83 (0.58) & \textbf{4.96} (0.21) & 4.50 (1.02) & \textbf{4.96} (0.21)\\
        2. Discomfort (1) & 1.08 (0.29) & \textbf{1.04} (0.21) & 1.50 (1.14) & 1.08 (0.42)\\
        3. Tension (1) & 1.08 (0.29) & 1.17 (0.49) & 1.50 (0.98) & \textbf{1.00} (0.00)\\
        4. Trust (5) & 4.29 (1.14) & \textbf{4.75} (0.62) & 4.17 (1.37) & 4.46 (1.20)\\
        5. Comfort (5) & NA & \textbf{4.54} (0.67) & 3.50 (1.44) & \textbf{4.54} (0.73)\\
        6. Convenience (5) & NA & \textbf{4.67} (0.57) & 3.87 (1.54) & \textbf{4.67} (0.83)\\
        7. Naturalness (5) & NA & \textbf{3.50} (1.34) & 2.58 (1.67) & 3.33 (1.71)\\
        8. Total Perception (10) & 8.62 (1.88) & 8.46 (2.04) & 7.42 (2.80) & \textbf{8.92} (1.52)\\
        \bottomrule
    \end{tabular}
    \label{tab:likert}
\end{table*} 

As for the subjective perception, the results of the eight interview questions conducted after each condition trial are summarized on Table~\ref{tab:likert}. Questions 2) and 3) addressed negative emotions, thus, lower scores (closer to $1$) were more positive. Questions 5), 6) and 7) explicitly referred to physical contact condition and, therefore, are not applicable to NC. Among the conditions, FC achieved the highest overall rating with an average score of $8.9$ out of $10$, while LA received the lowest with $7.4$. Furthermore, LA consistently obtained the poorest score across all individual questions. NC in turn performed moderately well but never achieved the best score in any question.

The results of the oral interview with the short version of the PANAS to assess emotional perception from positive and negative dimensions are summarized in Table~\ref{tab:panas_mean_sd}. A slight tendency toward FC as the best condition can be identified while LA received the lowest mean score for the positive affect and HH the highest mean score for the negative affect. However, the statistical analysis showed no significant differences (Table~\ref{tab:panas_anova}).

\begin{table}
\centering
\caption{Mean and standard deviation (SD) of Positive Affect (PA) and Negative Affect (NA) scores (PANAS) for each contact condition}
\begin{tabular}{lcc}
\toprule
\textbf{Condition} & \textbf{Mean} & \textbf{SD} \\
\midrule
\multicolumn{3}{c}{\textbf{Positive Affect (PA)}} \\
\midrule
No Contact (NC) & 4.12 & 1.03 \\
Hand Holding (HH) & 4.19 & 0.85 \\
Linking Arms (LA) & 4.08 & 1.01 \\
Full Forearm Contact (FC) & \textbf{4.26} & 0.85 \\

\midrule
\multicolumn{3}{c}{\textbf{Negative Affect (NA)}} \\
\midrule
No Contact (NC) & 1.04 & 0.11 \\
Hand Holding (HH) & 1.06 & 0.16 \\
Linking Arms (LA) & 1.03 & 0.09 \\
Full Forearm Contact (FC) & \textbf{1.01} & 0.07 \\
\bottomrule
\end{tabular}
\label{tab:panas_mean_sd}
\end{table}

\begin{table}
\centering
\caption{Results of the repeated-measures ANOVA for Positive and Negative Affect (PANAS) (Greenhouse–Geisser correction due to violation of sphericity)}
\begin{tabular}{lcc}
\toprule
\textbf{Affect} & \textbf{F} & \textbf{p-value} \\
\midrule
Positive Affect & 0.65 & 0.53 \\
Negative Affect & 1.12 & 0.32 \\
\bottomrule
\end{tabular}
\label{tab:panas_anova}
\end{table}

The NARS questionnaire was administered before and after the experiment to assess potential differences in participants' overall perception of robots. Table~\ref{tab:nars_mean_sd} summarizes the mean and standard deviation of the three subscales before and after the robot interaction, showing a small tendency toward an increase, mainly for Scale 2: ``Negative Attitudes toward Social Influence of Robots''. However, the Wilcoxon signed-rank test revealed no statistically significant differences.

\begin{table}
    \centering
    \caption{Mean and standard deviation (SD) of NARS scores before and after the robot interaction}
    \begin{tabular}{lcc}
    \toprule
        \textbf{Subscale}  & \textbf{Before} & \textbf{After} \\
        \midrule
        \footnotemark[1]Scale 1  & 1.97 (0.84) & 1.99 (0.70) \\
        \footnotemark[2]Scale 2  & 2.77 (0.91) & 3.06 (0.94) \\
        \footnotemark[3]Scale 3  & 2.71 (0.78) & 2.84 (0.79) \\
        \bottomrule
    \end{tabular}
    \footnotetext[1]{Scale~1 - \textit{Negative Attitudes toward Situations and Interactions with Robots}}
    \footnotetext[2]{Scale~2 - \textit{Negative Attitudes toward Social Influence of Robots}}
    \footnotetext[3]{Scale~3 - \textit{Negative Attitudes toward Emotions in Interaction with Robots}}
    \label{tab:nars_mean_sd}
\end{table}

%\begin{table}
%\centering
%\caption{Mean and standard deviation (SD) of NARS scores before and after the robot interaction.}
%\begin{tabular}{lcc}
%\toprule
%\textbf{Subscale} & \textbf{Mean} & \textbf{SD} \\
%\midrule
%\multicolumn{3}{c}{\textbf{Before the interaction}} \\
%\midrule
%Scale 1 & 1.97 & 0.84 \\
%Scale 2 & 2.77 & 0.91 \\
%Scale 3 & 2.71 & 0.78 \\
%\midrule
%\multicolumn{3}{c}{\textbf{After the interaction}} \\
%\midrule
%Scale 1 & 1.99 & 0.705 \\
%Scale 2 & 3.06 & 0.94 \\
%Scale 3 & 2.84 & 0.79 \\
%\bottomrule
%\end{tabular}
%\label{tab:nars_mean_sd}
%\end{table}

Feedback on the study as a whole, and on the broader topic of robots in healthcare, was diverse. Several participants recognized the potential benefits of robotic assistance noting: ``I think that being cared for by robots will be very useful for people who need help.'',``I can imagine that appropriate support would be very beneficial for a patient and at the same time relieve the burden on caregivers.''
However, many participants expressed cautious optimism or critical reflection.
Comments included: ``It is and remains a machine. I can well imagine that one day they will be able to do good service'', ``The study shows that robots can be used effectively as aids in everyday domestic tasks. The question is whether they can replace the human soul when someone is looking for empathy.'' A few participants voiced clear reservations, such as: ``Robots are devices, machines, and technical aids programmed by humans. They have no will of their own, are soulless, insensitive, and lifeless. Like all technology, they can also be defective or malfunction.''
One particularly striking comment read simply: ``Robots that express human emotions are extremely unpleasant.'' All in all, many participants reported enjoying the study and could imagine robots becoming valuable aids in healthcare contexts. An in-depth analysis of the responses to the open-ended question is presented in our complementary paper \cite{eckstein2026individual}.

\section{Discussion}\label{ch:evaluation}

This study investigates how older adults respond to and perceive different robotic touch modes during guidance. The study was conducted with 24 healthy, older adults using the robot TIAGo Pro as walking guide along a straight, 10 m-long path in four different, meaningful contact conditions: no contact, contact through holding the robot’s wrist with hand; contact through linking arms with the robot; and contact through resting the forearm on the robots forearm link. The evaluation combined physiological, behavioral, and subjective measures to quantify contact intensity, trust and acceptance, levels of stress, and subjective perception. Contact forces were used to quantify contact intensity and the nature of the interaction. The average distance between robots and the participant was associated with the intensity of the interaction and served as behavioral indicator of trust and confidence of the person toward the robot. ECG and EDA data captured involuntary physiological responses connected to levels of stress. Questionnaires in turn where used to capture the overall subjective perception. 

In contrast to the study by Piezzo and Suzuki \cite{piezzo_design_2016}, in which the participants could decide if they want to touch the robot, this study focused directly on robotic touch, with each participant participating in all four contact conditions. In addition, to obtain meaningful results, the number of participants is significant larger ($24$). The robot Pepper, used in \cite{piezzo_design_2016}, with its reduced height of $1.20$ m could also hinder the participants to walk with it side-by-side, thus in this study the humanoid robot TIAGo Pro with its adjustable height from $1.20$~m to $1.55$~m was employed. Furthermore, its long arms can reach up to $1.18$~m on both sides, enabling the implementation of comfortable poses.

No significant effects in \textit{HR} or \textit{HRV} have also been reported in other studies investigating robotic touch \cite{willemse_social_2019, guo_how_2024}.
One notable difference between those studies and the present results lies in the substantially different mean \textit{HRV} values ($14.5$ ms vs $31.5$ ms), which can be attributed to differences in participant demographics. This study focused exclusively on older adults, who are known to exhibit lower \textit{HRV} compared to younger populations \citep{umetani_twenty-four_1998}. In this study the individual conditions were not compared in terms of \textit{HR} or \textit{HRV}, but this could be explored in the future. 

The results of the EDA data suggest that participants experienced marginally lower stress levels when no physical contact with the robot occurred. However, since these effects are small, the conclusions should be interpreted with caution. It should be noted that during the physical contact conditions HH and FC, the EDA electrodes and the hand were in contact with the warm surface of the robot, which may have influenced the measurements.

The observed distribution of forces among the conditions was expected as the FC naturally involved the greatest contact area and thus the highest exerted force, whereas no force transfer was possible in NC. LA  exhibited the lowest median force among all contact conditions, which may be explained by the inconvenient posture reported by several participants, who were often unsure where to place their arm and often only simulated contact without fully touching the robot. HH also produced lower contact forces than the FC but higher than the LA, as it represents a more natural form of contact but involves a smaller contact area. However, overall the forces did not achieve high values, having a maximum value at approximately $20$~N. This suggests that participants perceived the robot more as a companion than a support device. That is possibly because all participants were able to walk independently, so they were not in need for physical assistance. Nevertheless, many seemed comfortable resting their arm on the robot in FC rather than avoiding contact or merely simulating it.

Spacial proximity was adopted as an objective behavioral indicator of how much the participants trust the robot. We assumed that a smaller relative distance between the robot and the participant reflects higher trust and confidence in the robot. The results suggest that the participants felt more uncomfortable in the LA condition, likely because they were required to stay closer to the robot, as the distance interval was small and confined near the robot (see Figure~\ref{fig:distance}). In fact, in LA the participants chose to walk farther from the robot within the available limited range, i.e.\ closer to the upper distance limit, with normalized (relative) distances closer to $1$, indicating distrust and emphasizing the discomfort with the imposed proximity in this contact mode. Interestingly, in NC, where participants could move as far away as the hallway allowed, participants still maintained relatively close distances both in absolute and relative terms, pointing to the probable existence of a range of comfortable distances. For the normalized value, however, there is no clearly defined upper limit for NC, as participants could theoretically walk as far as they wanted, which makes comparison of relative distances more difficult. For the analysis, the upper limit was arbitrarily set at approximately the distance to the lateral tape ($1.5$ m), even though participants were not told that this was a limit.

Regarding the subjective feedback, HH and FC received the top rating on five and six of the eight questions, respectively, suggesting that these two contact forms, corresponding to higher contact and interaction intensity, were perceived most positively overall. HH had better ratings in terms of discomfort, trust, and all three contact-related questions, indicating that this pose felt more natural and familiar to participants. In contrast, FC represents a less typical form of human-human contact during guidance, but suggests more safety and less tension, meaning the participants felt more relaxed. This interpretation aligns with participants' verbal feedback during the interview sessions. They often emphasized that this condition was enjoyable and they felt secure due to the larger contact area. 

Feedback for HH and NC was more mixed. While several participants also enjoyed these conditions, others perceived them as too slow, likely because in these scenarios they viewed the robot more as a companion than as a walking support, making the reduced speed seem less appropriate. After NC, some of the participants also said that they prefer a contact-based guiding. LA, in turn, received the most critical feedback, with participants describing it as uncomfortable and too rigid. Indeed, in comparison to connecting arms with humans, the robot's arm is devoid of soft tissue, with hard plastic covers instead. Furthermore, the robot's joints lack the compliance present in human arms. Interestingly, many participants commented positively on the ``bodily warmth'' of the robot. Although this effect was unintended, resulting from heat dissipation from the robot's motors, it nevertheless contributed to a more human-like and comforting impression.

Comparing the attitude toward robots in this population to the results of the NARS questionnaire applied in the study of Zhou et al.\ \cite{zhou_tactile_2021}, a difference in the mean values can be observed. While they reached an overall mean of $3.53$ in the first NARS questionnaire, in this study an overall mean of $2.54$ was obtained. The population in this study consisted of mainly female older adults with an average age of $75.8$~years, while the population in their study consisted of $13$ female and $8$ male participants, probably students, as the study took place at universities, with an average age of $24.7$ years. This indicates that older, female-biased participant pools may have a less negative attitude toward robots compared to the younger counterpart. This difference may be attributed to the younger population's greater exposure to social media content, showing video clips of robots out of control or revealing security gaps, which could foster a more cautious attitude toward robotic technology. Additionally, it can be speculated that concerns about robots potentially replacing human jobs may be more prevalent among younger individuals.

As a response to the open-ended question, several participants recognized the potential benefit of robots as aids in healthcare contexts, and many reported having enjoyed the study. At the same time, their critical reflections highlight a balanced perspective. While robots present promising opportunities for assistance and relief in caregiving, they also raise questions and concerns that warrant careful consideration. A deeper analysis of the open answers is discussed in the related paper \cite{eckstein2026individual}.

Finally, participants were invited to provide general feedback on the study and the robot itself. Regarding the robot, several participants suggested improvements to its appearance, such as making it more colorful, adding clothing, or equipping it with a more human-like hand. Others proposed that a softer or cushioned arm would enhance comfort during contact. A few participants also expressed the desire for verbal interaction during walking, which they felt would make the robot appear more natural. Additionally, some participants wished for greater control and flexibility, for example, the ability to guide the robot themselves or to adjust its walking speed.

The study has several limitations. Participants were mainly women already interested in research, and the experiment took place under controlled laboratory conditions with a teleoperated robot and short walking paths. Real-world trials with daily hospital routines could yield more ecologically valid results and capture interactions with a broader, more diverse population.

\subsection{Conclusion}
\label{se:conclusions}

This study investigated how older adults perceive and respond to robotic touch during walking guidance along an idealized straight path, combining physiological, behavioral, and subjective measures. ECG and EDA data captured involuntary physiological responses indicating stress levels. The magnitude of contact force indicated the contact intensity and the interaction nature, while distance served as behavioral indicators of trust and interaction intensity.  Questionnaires were used to assess subjective perception.  

Physiological results revealed a slight, non-significant increase in stress levels during robot interaction, which may partly stem from increased physical activity rather than from the contact itself. Similarly, a small rise in EDA during contact-based conditions suggests a cautious attitude toward close robot proximity, although this might also be influenced by the warmth of the robot's surface. Behavioral and subjective measures, however, point toward overall acceptance of robotic touch. Participants appeared more comfortable and trusting with the more contact and interaction intensive contact conditions, represented by placing the hand on the robots wrist link (HH), and placing the forearms on the horizontal forearm link of the robot (FC). FC provided stable support and a sense of safety, whereas HH was perceived as more natural and familiar. In contrast, the linking arm condition (LA) was least preferred, associated with reduced contact forces and larger normalized distances. These findings offer valuable insights for the design of assistive robots in geriatric care indicating that gentle, stable touch is appreciated by older adults, potentially fostering trust and comfort in this population, even if it elicits mild physiological arousal. Participants valued the sense of support and companionship provided by the robot and expressed interest in more human-like and communicative designs. Future work should focus on autonomous robot control that adapts the users via force and distance feedback. Integrating conversational capabilities or social cues could further enhance the perceived naturalness of interaction.

\backmatter

%\bmhead{Supplementary information} Not Applicable

%The supplementary material includes the data used for the evaluation:\\
%hr\_hrv\_pnn\_means.csv\\
%EDA\_all\_persons\_phases\_baseline\_gait.csv\\
%EDA\_all\_persons\_phases\_baseline\_sitting.csv\\
%force\_summary\_filtered\_with\_position.csv\\
%force\_summary\_conditions\_both\_arms.csv\\
%endeffector\_forces\_left\_5\_link.csv\\
%endeffector\_forces\_left\_6\_link.csv\\
%endeffector\_forces\_left\_7\_link.csv\\
%endeffector\_forces\_right\_5\_link.csv\\
%endeffector\_forces\_right\_6\_link.csv\\
%endeffector\_forces\_right\_7\_link.csv\\
%distance\_means.csv\\
%ecg\_individual\_traces.png\\

%The supplementary material includes the
%The German questionnaires?

%\bmhead{Acknowledgements}

\section*{Declarations}
\textbf{Funding} This work was made possible thanks to the endowment of the Chair ``Optimization \& Biomechanics for Human-Centred Robotics'' by the Hector Foundation II and the project funding by the Heidelberg Karlsruhe Strategic Partnership (HEIKA). \\
\textbf{Conflict of Interest} The authors declare that they have no conflict of interest. \\
\textbf{Ethical Standard} All participants were healthy older adults who were informed that if they so wished, they could withdraw from the experiment at any time.
Ethics approval for this study, which is part of the project “Perception of Robotic Touch in Geriatric Healthcare (PerRoT-G)”, ref. No. S-290/2024, was obtained from the Ethics Commission of the Medical Faculty, Heidelberg University.\\
\textbf{Informed Consent} Informed consent was obtained from all individual participants included in the study.\\
\textbf{Data availability} The descriptive statistics
are provided in Tables in the manuscript. 
All data on computed metrics for all participants and conditions used for the statistical analyses can be provided 
%within the Supplementary material in the form of spreadsheets. 
 by the corresponding author upon reasonable request.\\
\textbf{Materials availability} Not applicable \\
\textbf{Code availability} Not applicable\\
\textbf{Author contribution} LL, MA, CW, ME and KM designed the study. KM and ME secured funding and supervised the research. LL, TB and MS recruited participants and performed the experimental data acquisition. LL performed robot programming, data treatment, and statistical analysis. LL and MA wrote the first drafts of the manuscript, and all authors provided comments and revised all subsequent versions.\\
%\textbf{Open Access} Not applicable

\noindent
%If any of the sections are not relevant to your manuscript, please include the heading and write `Not applicable' for that section. 

%\bigskip

%\bigskip\noindent
%Springer journals and proceedings: \url{https://www.springer.com/gp/editorial-policies}

%\bigskip\noindent
%Nature Portfolio journals: \url{https://www.nature.com/nature-research/editorial-policies}

%\bigskip\noindent
%\textit{Scientific Reports}: \url{https://www.nature.com/srep/journal-policies/editorial-policies}

%\bigskip\noindent
%BMC journals: \url{https://www.biomedcentral.com/getpublished/editorial-policies}
%\end{flushleft}

\begin{appendices}

\section{Computation of Distance}\label{app:distance}

To determine the participant’s distance from the robot, first the DBSCAN (Density-Based Spatial Clustering of Applications with Noise) algorithm introduced by Ester et al. \cite{ester_density-based_1996} was used to cluster the points based on Euclidean distance. To simplify tracking, each cluster is represented as an axis-aligned bounding box (AABB) relative to the robot's frame, with the $x$-axis pointing forward, and the $y$-axis pointing medio-laterally. Each AABB is defined by its center and spatial extent. However, with only two dimensions, it is not possible to identify the human. For this reason, initially the participants stand in front of the robot so that the RGB-D camera could recognize them. Applying the body tracking algorithm to the data, the distance information of the human's torso frame was used to find the corresponding AABB from the laser scanner's point cloud. This AABB is then tracked over time, by identifying the closest AABB in the next iteration, and the participant’s distance and angle relative to the robot are computed from the center coordinates ($x$, $y$) of the AABB. This center could correspond either to the midpoint between the two legs, approximating the participant's overall center, or to the center of a single leg, if the other leg was occluded by the first leg or otherwise not detected by the laser scanners.




\end{appendices}


\bibliography{sn-bibliography.bib}% common bib file
%%%\input sn-article.bbl

\end{document}